\documentclass{article}

\PassOptionsToPackage{numbers, compress}{natbib}
 
\usepackage[preprint]{neurips_ai4d3_2023}

\usepackage[utf8]{inputenc} 
\usepackage[T1]{fontenc}    
\usepackage{hyperref}       
\usepackage{url}            
\usepackage{booktabs}       
\usepackage{amsfonts}       
\usepackage{nicefrac}       
\usepackage{microtype}      
\usepackage{xcolor}         
\usepackage{amsmath}
\usepackage{graphicx}
\usepackage{algorithm}
\usepackage{algpseudocode}

\title{FragXsiteDTI: Revealing Responsible Segments in Drug-Target Interaction with Transformer-Driven Interpretation}

\author{%
  Ali Khodabandeh Yalabadi\thanks{These authors contributed equally.} \\
  University of Central Florida\\
  Orlando, FL 32816 \\
  \texttt{yalabadi@ucf.edu} \\
  \And 
  Mehdi Yazdani-Jahromi\footnotemark[1] \\
  University of Central Florida\\
  Orlando, FL 32816 \\
  \texttt{yazdani@ucf.edu} \\
  \AND 
  Niloofar Yousefi \\
  University of Central Florida\\
  Orlando, FL 32816 \\
  \texttt{niloofar.yousefi@ucf.edu} \\
  \And 
  Aida Tayebi \\
  University of Central Florida\\
  Orlando, FL 32816 \\
  \texttt{aida.tayebi@ucf.edu} \\
  \And 
  Sina Abdidizaji \\
  University of Central Florida\\
  Orlando, FL 32816 \\
  \texttt{sina.abdidizaji@ucf.edu} \\
  \And 
  Ozlem Ozmen Garibay \\
  University of Central Florida\\
  Orlando, FL 32816 \\
  \texttt{ozlem@ucf.edu} \\
}

\begin{document}

\maketitle

\begin{abstract}
  Drug-Target Interaction (DTI) prediction is vital for drug discovery, yet challenges persist in achieving model interpretability and optimizing performance. We propose a novel transformer-based model, FragXsiteDTI, that aims to address these challenges in DTI prediction. Notably, FragXsiteDTI is the first DTI model to simultaneously leverage drug molecule fragments and protein pockets. Our information-rich representations for both proteins and drugs offer a detailed perspective on their interaction. Inspired by the Perceiver IO framework, our model features a learnable latent array, initially interacting with protein binding site embeddings using cross-attention and later refined through self-attention and used as a query to the drug fragments in the drug's cross-attention transformer block. This learnable query array serves as a mediator and enables seamless information translation, preserving critical nuances in drug-protein interactions. Our computational results on three benchmarking datasets demonstrate the superior predictive power of our model over several state-of-the-art models. We also show the interpretability of our model in terms of the critical components of both target proteins and drug molecules within drug-target pairs.
\end{abstract}

\newpage
\section{Introduction}
Drug–target interaction (DTI), representing the binding relationship between a drug and its target, is pivotal for developing new drugs and/or repurposing existing ones. While computational approaches have been present for several decades, their increased effectiveness and prominence as alternatives to High-Throughput Screening (HTS) have mainly been realized with the advancements of Machine Learning (ML) algorithms, with deep learning models offering a significant improvement in the accuracy of DTI predictions. Deep Learning's advantage lies in its ability to automatically capture valuable latent features, enabling it to handle intricate patterns in molecular data effectively. 


However, a notable constraint in utilizing deep learning models for drug discovery lies in the inherent lack of interpretability. While these models are powerful in predicting potential drug candidates, they often fall short in providing meaningful insights into why a particular compound exhibits certain properties or behaviors \cite{tang2023deep}.  This challenge pertains to understanding the intricacies of communication between proteins and drugs, each characterized by its own unique language or representation. Understanding the mechanistic basis of drug efficacy or inefficacy is critical for optimizing drug design \cite{yang2022deep}, refining candidate selection \cite{baptista2022evaluating}, and anticipating potential side effects or unforeseen interactions \cite{preto2021synpred}, all of which are fundamental to the drug development process. 

To address this challenge and further enhance the performance of deep learning models in drug discovery, we need \textbf{information-rich representations} of both proteins and drugs. These representations should encapsulate the structural, chemical, and functional aspects of each component comprehensively. For proteins, this may involve encoding details about their 3D structures, amino acid sequence, and binding sites. Similarly, it may entail capturing information about molecular fragments, chemical properties, and pharmacological characteristics of drugs. Another crucial component is a \textbf{mediator that serves as a common language}, bridging the gap between the two linguistic worlds of drug and protein. Such a common language facilitates information translation from the protein language to the drug language and vice versa. This translation process enables the model to effectively understand and interpret the interactions between proteins and drugs. This ensures that the information conveyed by proteins and drugs is not lost in translation, allowing the model to capture the nuances of their interaction, leading to improved interpretability and performance in DTI prediction tasks.   

Considering these challenges and leveraging the recent advancements in transformer-based models, we present FragXsiteDTI, an innovative transformer-based model that takes a new perspective on Drug-Target Interaction (DTI) prediction. Our approach provides a promising solution to both challenges by incorporating information-rich representations for both proteins and drugs alongside the integration of a learnable mediator that seamlessly connects these two distinct domains of drugs and proteins.

Our approach hinges on utilizing molecule fragments and protein pockets as the primary inputs to the model, a paradigm shift that enhances our understanding of the intricate interplay between drugs and their target proteins. Limited prior studies in the field have predominantly focused on using either protein pockets \cite{yazdani2022attentionsitedti} or drug fragments  for their DTI prediction models. However, to our knowledge, no existing research has harnessed protein pockets and drug fragments simultaneously as inputs in the context of DTI prediction. This finer granularity allows for a more precise analysis of which parts of the drug are critical for binding to specific binding sites of the protein targets. This information is pivotal in targeted and rationale drug design and generation. Instead of designing drugs as whole molecules and hoping to achieve effective interaction with a target protein, scientists can strategically design, combine or modify these fragments to optimize their ability to interact with specific proteins. Additionally, this detailed perspective goes beyond predictions to explanation by delving into the underlying mechanism of the casual-effect relationship.  Fragments often represent functional groups or motifs within drugs that directly contribute to their pharmacological activity. This granular-level analysis can help answer fundamental questions, such as why particular functional groups or binding regions are critical for interaction or why some drug-protein pairs do not exhibit the desired interaction. 

We introduce a transformer-based architecture inspired by the Perceiver IO framework, which facilitates the use of a mediator, enabling seamless communication between the distinct linguistic realms of proteins and drugs. This mediator is indeed a learnable latent array that undergoes a dynamic learning process. Initially, it is shaped based on the unique characteristics of proteins, allowing the model to focus on critical binding pockets of the protein. As the process unfolds, this query array undergoes adjustments influenced by the self-attention block. Then, our end-to-end learning process allows the model to fine-tune its focus, aligning the latent query with essential drug-related information. This learnable latent query array, guided by both proteins and drugs, is at the heart of our model's ability to decipher intricate drug-protein interactions effectively.

\begin{figure}[ht]
  \centering
  \includegraphics[width=1\textwidth]{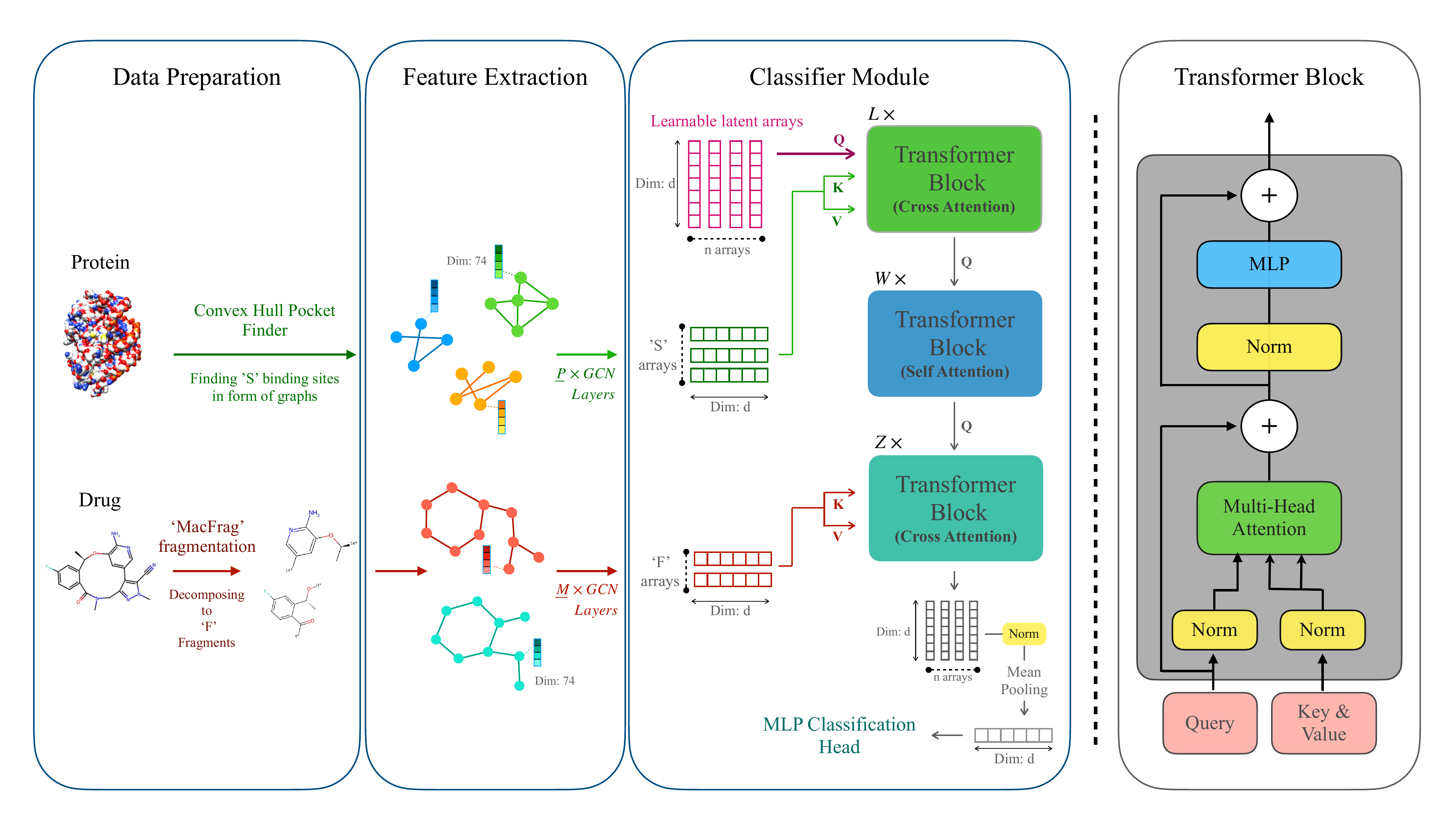}
  \caption{Our proposed framework, FragXsiteDTI, includes three main modules: (1) Preprocessing module, which consists of finding the binding sites of proteins, decomposing drugs' molecules into smaller fragments, and passing all of them to the next module in the form of graphs; (2) Feature extraction module, where we get graph representations of drugs' fragments and proteins' binding sites, and we create two multiple-layer graph convolutional neural networks to extract learnable embeddings; (3) Classifier module, where we introduce learnable latent arrays to first, find the most probable proteins' binding site(s) for interaction in a cross-attention transformer block(TB), second, pass through a self-attention TB to be prepared for finding the most probable drugs' fragment(s) in last cross-attention TB -  Transformer Block (TB) at the right shows the details of TBs in our classifier module.}
  \label{fig:model}
\end{figure}

The proposed framework can be found in Figure \ref{fig:model}.
The computational results on two datasets demonstrate the predictive power of our FragXsiteDTI compared to several state-of-the-art models and across multiple evaluation metrics. Also, our model is the first and the only one providing an information-rich interpretation of the interaction in terms of the critical parts of the target protein and drug molecule in a drug-target pair.

\section{Related works}
\label{related}

Deep learning-based approaches have emerged as effective solutions for addressing the challenging problem of Drug-Target Interaction (DTI) prediction. These approaches exhibit variations in both their architectural design and their strategies for representing input data.
\\
In DeepDTA \cite{ozturk2018deepdta}, a Convolutional Neural Network (CNN) is employed to analyze both the raw SMILES string and the protein sequence, allowing for the extraction of local residue patterns. The primary objective is to predict binding affinity values. To transform this into a binary classification problem, a Sigmoid activation function can be introduced at the end of the model.
DeepConv-DTI \cite{lee2019deepconv} employs CNN along with a global max-pooling layer to capture local patterns of varying lengths within the protein sequence. Additionally, it applies a fully connected layer to the drug fingerprint Extended-Connectivity Fingerprint 4 (ECFP4).
\\
Notably, while small drug molecules can be efficiently represented in one-dimensional space, proteins, due to their larger size and intricate interactions, often necessitate more comprehensive 3D representations.
Despite the limited availability of datasets containing 3D protein structures, recent deep learning literature has increasingly incorporated these structures into their investigations. For instance, AtomNet \cite{wallach2015atomnet} pioneered the utilization of 3D protein structures as input for a 3D Convolutional Neural Network (CNN), enabling the prediction of drug-target binding using a binary classifier. Ragoza et al. \cite{ragoza2017protein} proposed a CNN scoring function that harnessed the 3D representation of protein-ligand complexes to discern critical features crucial for binding prediction, surpassing the performance of the AutoDock Vina score. Pafnucy \cite{stepniewska2018development}, another significant advancement, employed 3D CNNs to predict binding affinity values for drug-target pairs. Their approach involved representing inputs as a 3D grid, treating both protein and ligand atoms similarly. By applying regularization techniques, their designed network focused on capturing the general properties governing interactions between proteins and ligands.
\\
These studies face several limitations, primarily stemming from the considerable difficulty in experimentally acquiring high-quality 3D protein structures. Consequently, there is a scarcity of datasets containing 3D structural information \cite{zheng2020predicting}. Moreover, most studies that employ 3D structural data predominantly rely on CNNs, which exhibit sensitivity to structural orientations and are computationally demanding.
\\
CPI-GNN \cite{tsubaki2019compound}, and GraphDTA \cite{nguyen2021graphdta} employ graph convolutional network (GCN) for molecular graph representation of drugs, and improve the prediction accuracy in the result. Some other studies \cite{gomes2017atomic,karimi2019deepaffinity} introduced the use of GCN approaches to take advantage of protein 3D structures as input for DTI prediction.
Some studies have extended the application of GCNs to protein-ligand complexes. A notable example is GraphBAR \cite{son2021development}, which stands as the pioneering 3D graph CNN utilizing a regression approach for predicting drug-target binding affinities. Instead of relying on 3D voxelized grids, GraphBAR employs graphs to represent protein-ligand complexes. These graphs manifest in the form of multiple adjacency matrices, with entries calculated based on distances and feature matrices encapsulating molecular properties of atoms. Additionally, GraphBAR augments its model with data derived from docking simulations.
\\
Lim et al. \cite{lim2019predicting} introduced a graph convolutional network model complemented by a distance-aware graph attention mechanism. This model extracts interaction features directly from the 3D structures of drug-target complexes generated through docking software. While their model demonstrated improved performance compared to both docking simulations and various deep learning-based models, it exhibited limitations, including reduced explainability and the introduction of additional docking errors into the deep learning model.
\\
Pocket Feature, proposed as an unsupervised autoencoder model by Torng et al. \cite{pocketgnn}, specializes in learning representations from binding sites within target proteins. The model employs 3D graph representations for protein pockets and 2D graph representations for drugs. It trains a GCN model to extract features from these graph representations and drug SMILEs. Notably, Pocket Feature outperforms 3D CNNs, as demonstrated in reference \cite{ragoza2017protein}, and also surpasses docking simulation models such as AutoDock Vina \cite{viana}, RF-Score \cite{humanbase}, and NNScore \cite{humanbase}.
\\
Zheng et al. \cite{zheng2020predicting} drew attention to the inefficiency of employing direct 3D structure inputs. Instead, they adopted 2D distance maps to represent proteins. In doing so, they reformulated the DTI prediction problem as a classical visual question-answering (VQA) task. In this paradigm, given a distance map of a protein, the model determines whether a given drug interacts with the target protein. Although their model outperformed several state-of-the-art models, it was primarily tailored to classification, predicting interactions between drug-target pairs, without any interpretability.
\\
With the emergence of Transformers and their proven power, TransformerCPI \cite{chen2020transformercpi} utilizes sequence representation of drugs and proteins. After a customized encoder layer, this model employs a transformer decoder to construct an interaction sequence and predict the interaction based on that.
MolTrans \cite{huang2021moltrans} combines the transformer's capability with sub-structures of drugs and targets. This model decomposes drugs and proteins into smaller structures based on their graphs and sequences respectively. Then, an interaction module is responsible for explicable prediction.
\\
In more recent studies, Yazdani et al. \cite{yazdani2022attentionsitedti} exploited 3D protein binding sites along with graph attention embeddings for both drug and protein. Their model finds protein binding sites by a simple docking-based model proposed by Fathi et al. \cite{fathi2014}. By using a self-attention module and sentence-level relation from NLP literature, this model determines the most probable candidate binding site for interaction resulting in interpretability as well as high accuracy. Furthermore, they demonstrated that the 3D structure of protein binding sites carries an immense amount of information related to binding affinity and it can improve the performance of benchmark models just by adding this information to the classifier \cite{bindingsiteaugmenteddta}.
\\
CSDTI \cite{pan2023csdti} is another model that focuses on the representation of drugs and proteins to improve interpretability and gain better performance in prediction. They employed a drug molecule aggregator and the multiscale 1D convolution-based protein encoder in order to extract the best representation and a cross-attention block to learn the relation between them. Focusing on representation, the MDL-CPI model \cite{wei2022mdl} considers proteins as words and employs a hybrid network architecture based on BERT and CNN to extract the feature representations. Following this work, AMMVF-DTI \cite{wang2023ammvf} is a model that utilizes local and global information of proteins and ligands in the form of node-level and graph-level embeddings, respectively. This model tried to improve performance with attention mechanisms and interactive information.

\section{Methodology}
\label{method}
Our architecture is delineated into three principal modules:

\begin{enumerate}
    \item \textbf{Data Preparation Module}: This module employs a fragmentation algorithm to dissect drugs and a simulation method to identify protein binding sites. Subsequently, a unique graph is constructed for each drug fragment and protein binding site. A comprehensive explanation is provided in Subsection \ref{dataprep}.
    
    \item \textbf{Feature Extraction}: Within this module, we utilize message-passing neural networks (MPNNs) to extract features from fragments and sites, encapsulating them into desired-sized embeddings. Further details are expounded in Subsection \ref{feature_extract}.
    
    \item \textbf{Classifier Module}:  Inspired by the Transformer and Perceiver IO architectures, our approach systematically models drug-protein interactions in a tripartite manner. Initially, a learnable query is employed to attend to protein embeddings. This query undergoes refinement and subsequently interacts with drug fragment embeddings, culminating in a holistic interaction representation. Detailed insights are presented in \ref{classifier}.
\end{enumerate}
A schematic representation of the entire architecture is illustrated in Figure~\ref{fig:model}.

\subsection{Data Preparation}
\label{dataprep}
\subsubsection{Extracting Fragments from The Ligand Molecule (MacFrag)}
Creating high-quality fragment libraries by dividing organic compounds is a crucial aspect of drug discovery. Utilizing fragments of drug molecules can result in a more information-rich representation embedding. In this regard, we chose a recent paper in this domain called Macfrag \cite{diao2023macfrag}. This study introduces a novel approach for efficiently fragmenting molecules. MacFrag employs an adapted version of BRICS rules to cleave chemical bonds and introduces an efficient algorithm for swiftly extracting subgraphs, enabling rapid enumeration of the fragment space. Using this approach, the size of fragments can vary depending on the number of chosen building blocks, enabling flexibility. This method permits overlaps between fragments, ensuring that no critical information is overlooked. Moreover, based on their experiments, the fragments generated using this approach exhibit a closer adherence to the 'Rule of Three.'

\subsubsection{Extracting Protein’s Ligand-binding Pockets}
We leverage the 3D configurations of proteins derived from the Protein Data Bank (PDB) files. PDB datasets comprise experimental measurements from Nuclear Magnetic Resonance (NMR), x-ray diffraction, cryo-electron microscopy, etc. associated with proteins. To identify protein binding sites, we employ the algorithm introduced by Saberi Fathi et al. \cite{fathi2014}. This approach stands out for its simplicity in extracting protein binding sites from their 3D configurations. It operates on a simulation-based paradigm and can be applied before feeding the data to a deep-learning architecture. It simplifies the deep-learning models because they are no longer responsible for processing large molecules of protein. The algorithm calculates bounding box coordinates for each protein binding site. These coordinates subsequently simplify the entire protein structure to a selection of peptide segments.

\subsubsection{Graph Construction}
Post the extraction of protein binding sites and ligand fragmentation, we devised a representation technique. For proteins, atoms in binding sites are nodes in distinct graphs, with edges defined by inter-atomic distances below a threshold. For ligand fragments, edges are determined by atomic bonds. Atom features, encoded via one-hot vectors, encompass atom type, degree, implicit valence, charge, radical electrons, hybridization, aromaticity, and attached hydrogen count, yielding a $1 \times 74$ vector per node (details can be found in the appendix). This methodology produces multiple graphs for both individual proteins and distinct ligands.

\subsection{Feature Extraction}
\label{feature_extract}
Subsequent to this Data Preparation stage, we employed message-passing neural networks (MPNNs) to encapsulate the constructed graphs into embeddings. This serves as the feature extraction component of our architecture. In the appendix, we delineate the descriptions and functionalities of the layers utilized.
We combined two layers of TAGCN followed by a GAT layer; the model harnesses the strengths of both methods, potentially outperforming architectures that rely solely on TAGCN or GAT. The TAGCN layers adeptly capture varying local structures by considering different powers of the adjacency matrix, ensuring sensitivity to the graph's topology and enhancing local feature extraction \cite{du2017topology}. Subsequently, the GAT layer introduces an attention-based pooling mechanism, allowing nodes to assign varying importance scores to their neighbors, emphasizing more relevant nodes and down-weighting less pertinent ones \cite{velivckovic2017graph}. This selective attention mechanism leads to more discriminative graph embeddings, especially in graphs with varying node importance. Moreover, the combination ensures that the model recognizes diverse local structures while discerning node importance, offering a dual capability beneficial in complex graphs. Additionally, the GAT layer can mitigate the over-smoothing problem often seen in deep graph neural networks, ensuring distinct and informative node representations \cite{li2018deeper}.

\subsection{Classifier Module}
\label{classifier}
We present a method to model drug-protein interactions, drawing inspiration from the Transformer architecture \cite{vaswani2017attention} and the Perceiver IO model \cite{jaegle2021perceiver}. Our approach is delineated into three distinct stages:

\begin{enumerate}
    \item \textbf{Cross-Attention with Learnable Query:} We initiate with a learnable latent query array. This query attends to protein binding site embeddings through a cross-attention mechanism. The outcome of this stage is a weighted representation of protein binding sites based on the learnable query.
    
    \item \textbf{Latent Query Processing:} The weighted representation from the first stage undergoes further refinement via a self-attention Transformer block. This processed query encapsulates the nuanced features of protein binding sites, preparing it for interaction modeling with drug fragments.
    
    \item \textbf{Drug-Protein Interaction Modeling:} In this stage, the processed query from the preceding step acts as $Q$ in a cross-attention Transformer block. Drug fragment embeddings serve as both $K$ and $V$. This setup allows the model to focus on drug fragments that are most pertinent in relation to the protein binding sites, yielding a holistic representation of drug-protein interaction dynamics.
\end{enumerate}
Our approach to modeling drug-protein interactions, while bearing foundational similarities with the Perceiver IO architecture \cite{jaegle2021perceiver}, diverges in its handling of the latent queries and the source of keys and values for attention. Both methodologies employ learnable latent queries to attend to input data, facilitating the extraction of intricate patterns without the need for domain-specific architectures. However, in our method, we introduce a unique twist after the initial cross-attention between the latent query and protein binding site embeddings and subsequent latent query processing. Instead of relying on another set of learnable queries or expert-generated ones for outputs, as in Perceiver IO, we source the keys and values directly from the drug fragment space. This processed query then interacts with the drug fragment embeddings in a cross-attention Transformer block, with the query as $Q$ and the drug fragment embeddings serving as both $K$ and $V$. This design choice tailors our approach to more effectively capture the nuances of drug-protein interactions.

\paragraph{Self-Attention Mechanism} assigns weights to different segments of an input sequence when formulating an output sequence. Defined as Equation \ref{eq:att}.
\begin{equation}
\label{eq:att}
    \text{Attention}(Q, K, V) = \text{softmax}\left(\frac{QK^T}{\sqrt{d_k}}\right) V 
\end{equation}

where $Q$, $K$, and $V$ are the query, key, and value matrices, respectively, and $d_k$ is their dimensionality. This mechanism adeptly captures long-range dependencies without the constraints of recurrent layers.

\paragraph{Cross-Attention Mechanism} the queries come from one sequence (or representation), while the keys and values come from another. This allows the model to focus on relevant parts of the second sequence based on the information from the first, effectively bridging information between two distinct sources.

\paragraph{Transformer Block}is a composite of attention layers, feed-forward networks, and normalization organized in a layered fashion.

\section{Experiments}
\label{experiments}

\subsection{Datasets}
We establish the effectiveness of our proposed model through a series of comparative experiments. In these experiments, we evaluate the performance of FragXsiteDTI alongside several state-of-the-art methods. To do so, we employ two benchmark datasets that offer essential 3D structural information of target proteins, a crucial requirement for our model.

\paragraph{Human, \textit{C.elegans}, and DrugBank}
These datasets were constructed by amalgamating a collection of exceptionally trustworthy and dependable negative drug-protein samples through a systematic in silico screening technique, which was then combined with the known positive samples from sources such as HumanBase \cite{humanbase} and the comprehensive DrugBank database \cite{drugbank}. The DrugBank dataset included 68696 positive and 67072 negative drug-protein interactions, with a total of 6641 unique drugs and 3547 unique targets, providing a robust foundation for interaction prediction models. The human dataset comprises 3,369 positive interactions involving 1,052 distinct compounds and 852 unique proteins; the \textit{C. elegans} dataset encompasses 4,000 positive interactions, encompassing 1,434 distinct compounds and 2,504 unique proteins.\\
We also, made a contribution to the enhancement of the DrugBank and \textit{C.elegans} datasets by augmenting the PDB (Protein Data Bank) IDs of the target sequences to the datasets. This means that we have successfully extracted the three-dimensional structures of the proteins contained within these datasets. By doing so, we have enriched the datasets, paving the way for future research and applications that rely on 3D structure-based methods. 
To enable a direct comparison, identical train, validation, and test partitions (80\%, 10\%, 10\%) were adopted, as in recent studies \cite{zheng2020predicting, yazdani2022attentionsitedti}.

\subsection{Implementation and evaluation}

\emph{Experimentation strategies}. For our implementations, we utilized PyTorch 1.8.2, a long-time support version. The appendix contains all the hyperparameters employed for our model. The experimentation was conducted on an Nvidia RTX 3090 GPU with 24 GB of memory.

\emph{Evaluation metrics}. We conducted evaluations of our models using various metrics, such as the Area Under the Receiver Operating Characteristic Curve (AUC), precision, recall, and the F1 score. 


\subsection{Comparison on target datasets}
\subsubsection{Human}
We conducted a comprehensive comparison of our model against recently developed deep learning-based approaches, including GraphDTA \citep{nguyen2021graphdta}, CPI–GNN \citep{tsubaki2019compound}, DrugVQA \citep{zheng2020predicting}, AttentionSiteDTI \citep{yazdani2022attentionsitedti}, DeepDTA \citep{ozturk2018deepdta}, DeepConv-DTI \citep{lee2019deepconv}, MolTrans \citep{huang2021moltrans}, TransformerCPI \citep{chen2020transformercpi}, as well as CSDTI \citep{pan2023csdti}. In Section \ref{related}, you can find a brief overview of each of these models.
\\
The performance of these models is summarized in Table \ref{tab:human}. Notably, our proposed model demonstrates superior prediction performance compared to all these models. It achieves competitive results with AttentionSiteDTI, which currently holds the top performance among them. Deep learning models prove to be highly effective in extracting essential features governing the intricate interactions within drug-target pairs. Building upon this foundation, our model further enhances accuracy, highlighting the quality of features extracted and learned during the end-to-end training process of fragXsiteDTI.

\begin{table*}[ht]
\centering
\caption{Human Dataset Comparision}
\label{tab:human}
\begin{tabular*}{\textwidth}{@{\extracolsep{\fill}}lcccc@{\extracolsep{\fill}}}
\toprule
& AUC           & Precision        & Recall          & F1 Score        \\ 
\midrule
GraphDTA     & 0.960           & 0.882            & 0.912           & 0.897           \\ 
GCN          & 0.956           & 0.862            & 0.928           & 0.894           \\ 
CPI–GNN      & 0.970           & 0.923            & 0.918           & 0.920           \\ 
DrugVQA      & 0.979           & 0.954            & 0.961           & 0.957           \\
DeepDTA	& 0.972 & 	0.938	& 0.935	& 0.936 \\
DeepConv-DTI &	0.967	&  0.939	&  0.907	&  0.923 \\
MolTrans &	0.974	& 0.955	 &  0.933	&  0.944 \\
TransformerCPI &	0.97	& 0.911	&  0.937	&  0.924 \\
AttentionSiteDTI & 0.991 & 0.951 & \textbf{0.975} & 0.963 \\
CSDTI	& 0.982 & 	0.937	& 0 .946	&  0.941 \\
AMMVF-DTI 	& 0.986	& 0.976	 & 0.938	& 0.957 \\
FragXsiteDTI(Ours)	& \textbf{0.991}	& \textbf{0.977}	& 0.952	& \textbf{0.964}\\
\bottomrule
\end{tabular*}
\end{table*}

\subsubsection{C.elegans}
For this dataset, we compared our model with deep learning models that had high performance based on the selected metrics. These models include MDL-CPI \cite{wei2022mdl}, CPI-GNN \cite{tsubaki2019compound}, graph convolutional network (GCN), GraphDTA \cite{nguyen2021graphdta}, TransformerCPI \cite{chen2020transformercpi}, and AMMVF-DTI \cite{wang2023ammvf}. The performance of these models is summarized in Table \ref{tab:c.elegans}. Similar to the Human dataset, our model outperforms the existing good models in all prediction metrics.

\begin{table*}[ht]
\centering
\caption{\textit{C.elegans} Dataset Comparision}
\label{tab:c.elegans}
\begin{tabular*}{\textwidth}{@{\extracolsep{\fill}}lcccc@{\extracolsep{\fill}}}
\toprule
                & AUC           & Precision        & Recall          & F1 Score        \\ 
\midrule
MDL-CPI	        & 0.975	           & 0.943	       & 0.923           & 0.933           \\
CPI-GNN	            & 0.978     	   & 0.938         & 0.929	         & 0.933           \\
GCN 	        & 0.975	           & 0.921	       & 0.927	         & 0.924           \\
GraphDTA 	    & 0.974	           & 0.927	       & 0.912	         & 0.919           \\
AttentionSiteDTI & 0.988            & \textbf{0.974}        &0.932   & 0.953 \\
TransformerCPI 	& 0.988	           & 0.952	       & 0.953	         & 0.952           \\
AMMVF-DTI 	    & 0.990	            & 0.962	       & 0.96	         & 0.961           \\
FragXsiteDTI(Ours)	& \textbf{0.992}	       &0.971	       & \textbf{0.971}	         & \textbf{0.971}           \\
\bottomrule
\end{tabular*}
\end{table*}

\subsection{DrugBank}
For the DrugBank dataset, we benchmarked our model against a suite of advanced computational models known for their high performance according to the chosen evaluation metrics. Specifically, we compared with methods such as Random Walk with Restart (RWR) \cite{lee2018identification}, DrugE-Rank \cite{yuan2016druge}, DeepCPI \cite{wan2019deepcpi}, DeepConv-DTI \cite{lee2019deepconv}, and MHSADTI \cite{cheng2021drug}. These methods represent a diverse range of approaches, from network-based algorithms like RWR and DrugE-Rank, which prioritize the global structure of drug-target interaction networks, to deep learning architectures such as DeepCPI and DeepConv-DTI, which leverage convolutional networks, and MHSADTI that integrates multiple heterogeneous sources for drug-target interaction prediction. The comparative performance analysis of these models is detailed in Table \ref{tab:drugbank}. Our results indicate that, akin to what was observed with the Human and \textit{C.elegans} datasets, our model surpasses the aforementioned well-established models across all prediction metrics.

\begin{table*}[ht]
\centering
\caption{Drugbank Dataset Comparision}
\label{tab:drugbank}
\begin{tabular*}{\textwidth}{@{\extracolsep{\fill}}lcccc@{\extracolsep{\fill}}}
\toprule
                & AUC           & Precision        & Recall          & F1 Score        \\ 
\midrule
RWR            & 0.759                       & 0.704                       & 0.651                       & 0.677 \\
DrugE-Rank     & 0.759                       & 0.707                        & 0.629                       & 0.666 \\
DeepConv-DTI   & 0.853                       & 0.789                       & 0.738                       & 0.763 \\
DeepCPI        & 0.700                       & 0.700                       & 0.556                       & 0.62  \\
MHSADTI        & 0.863                       & 0.771                       & 0.792                       & 0.781 \\
TransformerCPI & 0.865 & 0.774 & 0.821 & 0.797 \\
FragXsiteDTI(Ours)   & \textbf{0.901} & \textbf{0.816} & \textbf{0.842} & \textbf{0.829} \\
\bottomrule
\end{tabular*}
\end{table*}

\subsection{Interpretation}
\subsubsection{Protein}
\begin{figure}[ht]
  \centering
  \includegraphics[width=1\textwidth]{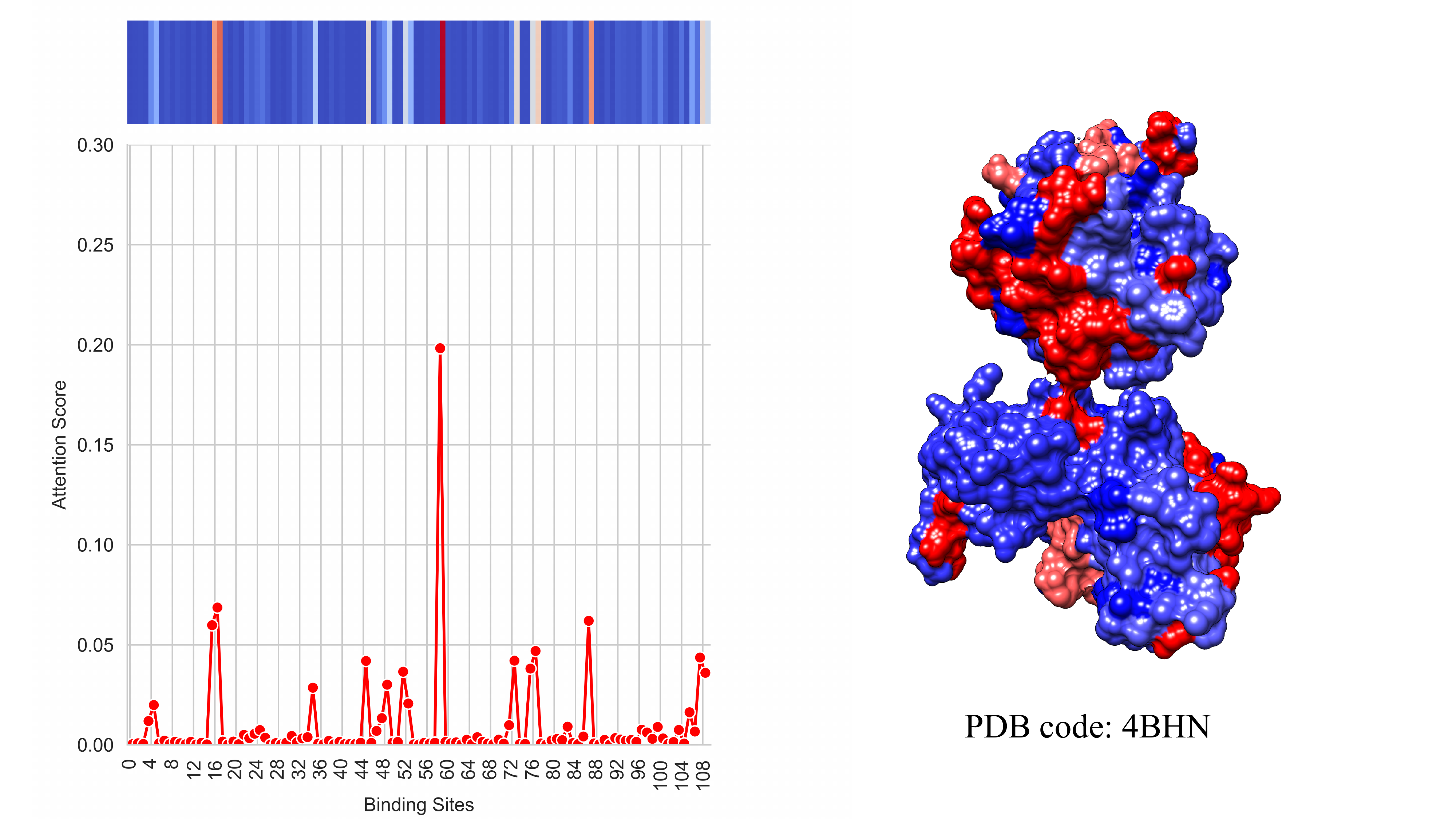}
  \caption{(Left) a heatmap and line plot that represent the cross-attention mechanism weights for every binding site with the latent array. These weights signify the likelihood of each computed binding site on the protein becoming active when interacting with the specific ligand ($C_{21}H_{23}Cl_{2}N_{5}O_{2}$). (Right) a heatmap that projects the cross-attention weights onto the protein with the PDB code 4BHN. This visualization illustrates our model's interpretability with respect to the proteins. The protein's visualization was generated using UCSF Chimera software \cite{pettersen2004ucsf}.}
  \label{fig:binding_heatmap}
\end{figure}

In this study, we leverage the attention mechanism to enhance the model's ability to predict the likelihood of specific protein binding sites interacting with a given ligand. This likelihood is quantified through the attention matrix computed within the model. The visualization of this attention mechanism can be observed in Figure \ref{fig:binding_heatmap}, where it is presented as a heatmap for the protein with PDB code of 4BHN when interacting with a drug characterized by the molecular formula of $C_{21}H_{23}Cl_{2}N_{5}O_{2}$. This figure also includes the projection of the heatmap onto the protein structure.

\subsubsection{Drug}
\begin{figure}[ht]
  \centering
  \includegraphics[width=1\textwidth]{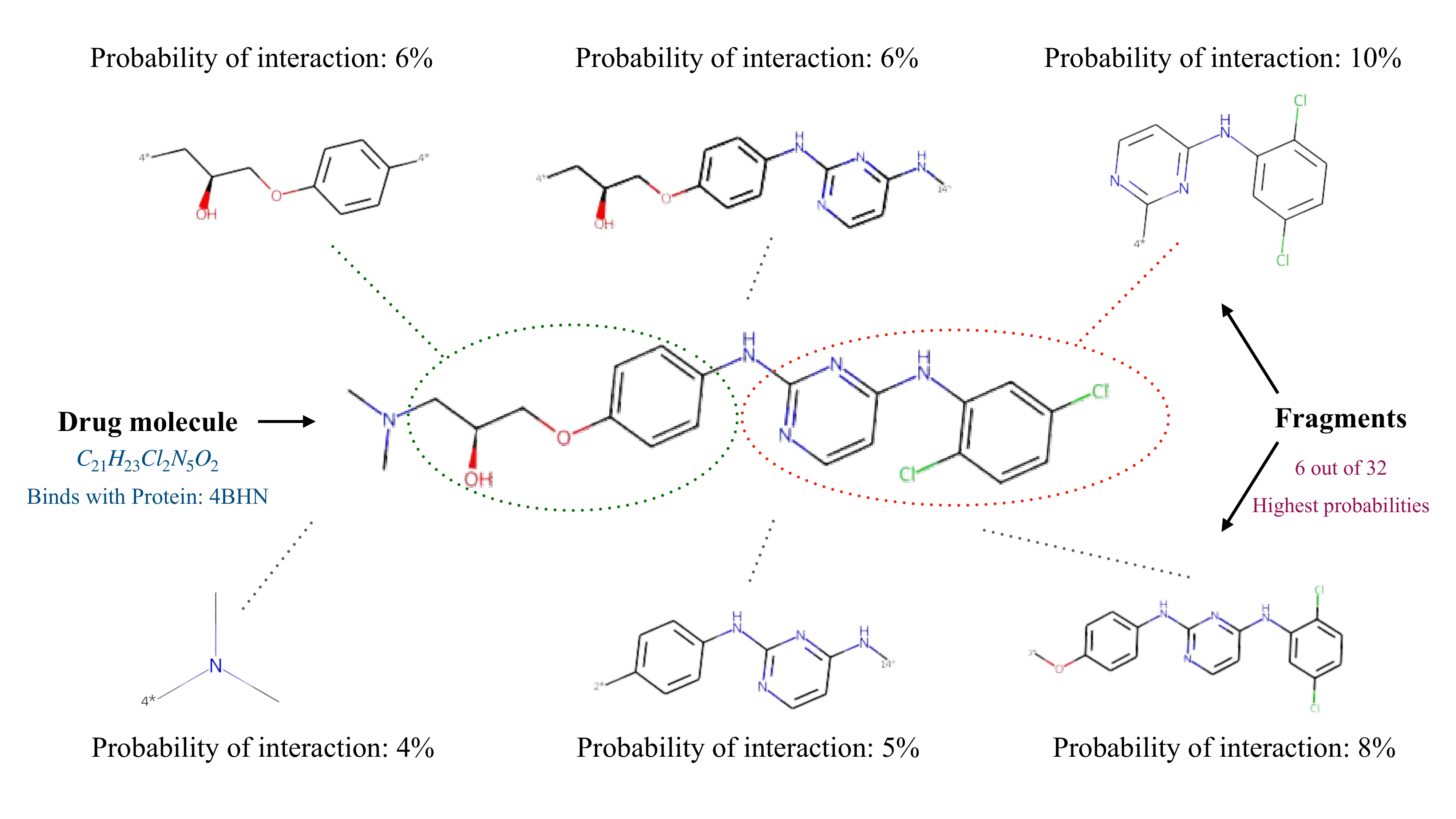}
  \caption{This figure demonstrates the interpretability of our model regarding the drug molecules. It showcases the specific drug molecule that binds to protein 4BHN, along with its six fragments exhibiting the highest interaction probabilities among a total of 32 fragments. These probabilities are normalized, and 10\% is the highest. Note that since these fragments can be repeated in other fragments, these probabilities can be summed up to higher numbers for each fragment.}
  \label{fig:drug}
\end{figure}

We also have an attention matrix for each drug molecule that determines which fragments of that drug have the highest probabilities for interaction with the particular protein. These candidate fragments can explain the chemical properties that caused the interaction or be used for designing and generating new drugs. Figure \ref{fig:drug} demonstrate an example of certain drug ($C_{21}H_{23}Cl_{2}N_{5}O_{2}$) that binds with the target protein (4BHN).

\begin{figure}[ht]
  \centering
  \includegraphics[width=1\textwidth]{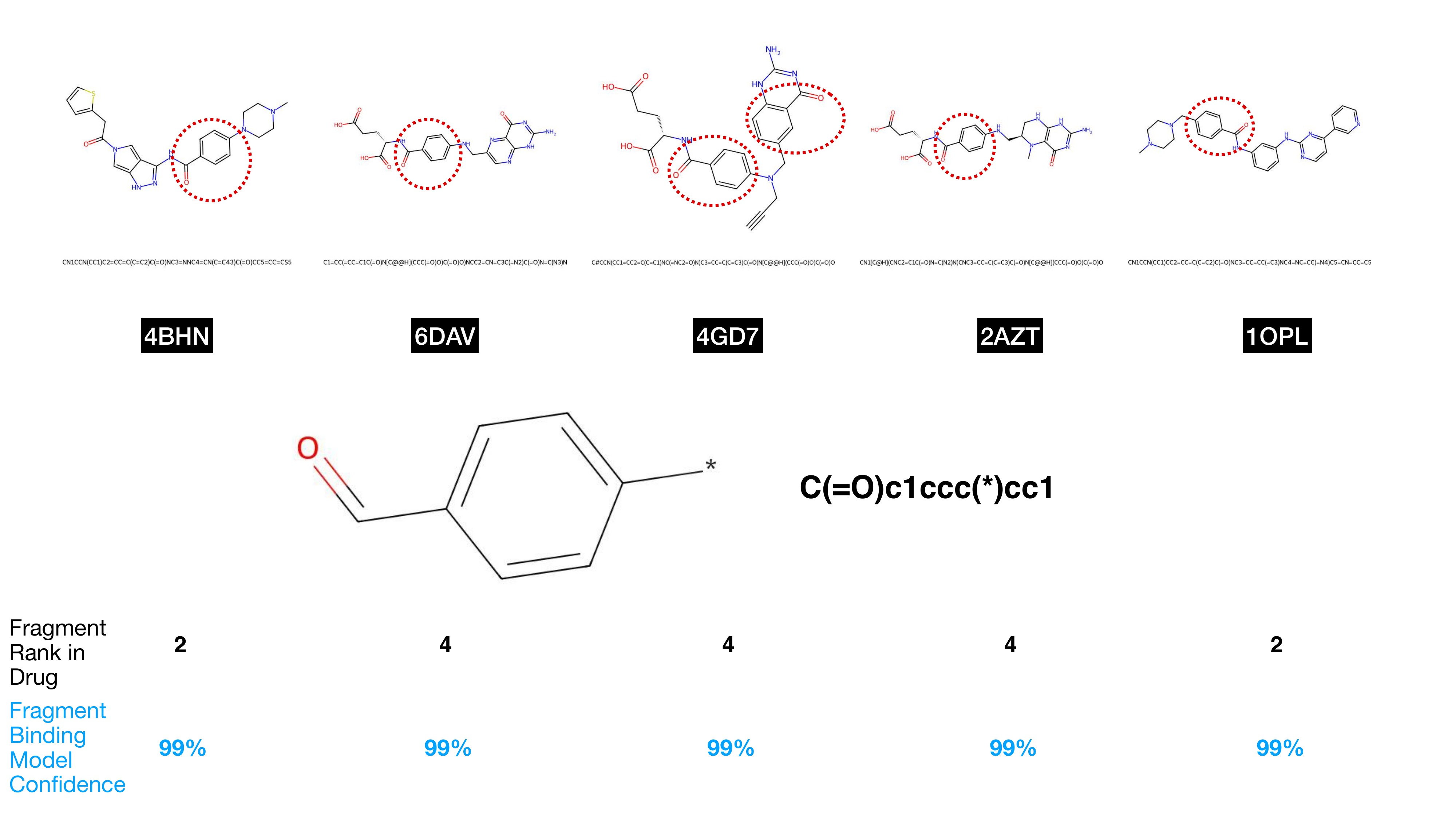}
  \caption{Visualization of key molecular fragments within various drugs, highlighted in red dotted circles, showcasing their potential significance in mediating drug-protein interactions. Consistent appearance of the same fragment across multiple drugs underscores its pivotal role in binding. Model-derived rankings and confidences emphasize the fragment's importance}
  \label{fig:drug_inter}
\end{figure}

The provided depiction of multiple interactions in figure \ref{fig:drug_inter} illustrates the structural interactions between specific drugs and their associated proteins, as determined by our advanced computational model. Central to the depicted interactions is the presence of distinct molecular fragments within each drug, highlighted by the dotted red circles. These fragments are of utmost importance as they represent the primary contact points or key residues that likely mediate the drug's binding affinity to its target protein.

For instance, for the drug associated with protein '4BHN', our model has ranked the highlighted fragment as the second most crucial component within the drug's structure for mediating protein binding. This rank is further supported by an impressive 99\% confidence level in the fragment's binding model. This trend of high-ranking fragments paired with high confidence levels is consistent across the displayed drugs, such as those paired with proteins '6DAV', '4GD7', '2AZT', and '10PL'. Specifically, the fragments in these drugs are ranked 2, 4, 4, and 2, respectively, all backed by a binding model confidence of 99\%.

An intriguing observation from the visualization is the recurring appearance of the same molecular fragment across different drugs. This consistency accentuates the significance of this particular fragment in mediating drug-protein interactions, suggesting its central role in the binding process.

Such findings hold transformative implications for the realm of drug discovery. By pinpointing and understanding these pivotal molecular fragments within drug compounds, researchers are presented with a novel approach to optimize drug design. Recognizing fragments responsible for effective binding can pave the way for the design of new drugs, where these fragments can be incorporated or modified to enhance binding affinity, selectivity, or other pharmacological properties. Essentially, by targeting and modifying these "hotspots", there's potential to not only improve existing drugs but also innovate the development of next-generation therapeutics.

However, it's imperative to highlight that these results are derived from computational analyses. While they offer insightful and potentially groundbreaking perspectives, they must be rigorously validated in laboratory settings to confirm their efficacy and accuracy. As with all computational findings, empirical evidence from experimental assays is crucial to establish the true potential of these identified drug fragments.

\section{Conclusion}
In this work, we introduced a groundbreaking method for modeling drug-protein interactions, drawing from the robustness of both the Transformer and Perceiver IO architectures. Our empirical evaluations on Human and \textit{C. elegans} datasets not only underscore the superiority of our approach in terms of predictive accuracy but also highlight its unparalleled interpretability. One of the standout features of our method is its capability to pinpoint which fragment of a drug interacts with specific regions of a protein. This granularity is invaluable, offering researchers a detailed map of interaction hotspots, which can guide drug modifications and optimizations. The use of attention modules doesn't merely serve as a mechanism for improved performance; it acts as a window into the model's decision-making process. By visualizing attention scores, we can discern the importance the model assigns to different fragments, shedding light on potential pharmacologically active regions. With the insights provided by our model, drug designers can make informed decisions, potentially reducing the trial-and-error nature of drug discovery. By knowing which fragments are likely to interact with target proteins, drug modifications can be more strategic and purpose-driven. Given the model's precision in understanding drug-protein dynamics, there's potential for tailoring drug designs to individual protein structures, paving the way for more personalized therapeutic interventions in the future. While our current evaluations are on specific datasets, the foundational architecture suggests potential scalability to other organisms and broader drug-protein interaction landscapes, making it a versatile tool in the bioinformatics toolkit. As the pharmaceutical industry and medical research communities continue their quest for more effective and targeted drugs, our method stands out as a beacon, promising to play a transformative role in the future landscape of drug discovery and design.

\paragraph{Data and Code availability} 
All datasets and all instructions and codes for our experiments are publicly available at \url{https://github.com/yazdanimehdi/FragXsiteDTI}

{
\small

\bibliographystyle{plainnat}
\bibliography{bibliography}
}


\end{document}